\documentclass[conference]{IEEEtran}
\IEEEoverridecommandlockouts
\usepackage{cite}
\usepackage{amsmath,amssymb,amsfonts}
\usepackage{algorithmic}
\usepackage{graphicx}
\usepackage{textcomp}
\usepackage{xcolor}
\usepackage{authblk}
\usepackage{booktabs}
\usepackage{hyperref}
\usepackage{ulem}
\usepackage[switch]{lineno}

\usepackage{acro}

\usepackage{xr}

\makeatletter
\newcommand*{\addFileDependency}[1]{
\typeout{(#1)}
\@addtofilelist{#1}
\IfFileExists{#1}{}{\typeout{No file #1.}}
}\makeatother

\DeclareAcronym{att}{
    short = sVA, 
    long = spiking visual attention
    }

\DeclareAcronym{sl}{
    short = sSLF, 
    long = spiking sign language 
}

\DeclareAcronym{mnist}{
    short = SL-MNIST, 
    long = Sign Language MNIST dataset
}

\DeclareAcronym{asl}{
    short = ASL-DVS, 
    long = ASL-DVS dataset
}

\DeclareAcronym{sr}{
    short = sR, 
    long = spiking recognition 
}

\def\BibTeX{{\rm B\kern-.05em{\sc i\kern-.025em b}\kern-.08em
    T\kern-.1667em\lower.7ex\hbox{E}\kern-.125emX}}

\begin{document}

\title{Neuromorphic visual attention for Sign-language recognition on SpiNNaker\\

}
\author[1*]{Šárka Lísková\thanks{$^*$These authors contributed equally. Corresponding authors: \texttt{sarka.liskova@fel.cvut.cz}, \texttt{vedmeolh@fit.cvut.cz}}}
\author[2*]{Olha Vedmedenko}
\author[2]{Mazdak Fatahi}
\author[1]{Matej Hoffmann}
\author[3]{P. Michael Furlong}
\author[1]{Giulia D'Angelo}

\affil[1]{Dept. of Cybernetics, Faculty of Electrical Engineering, Czech Technical University in Prague, Czech Republic}
\affil[2]{Faculty of Information Technology, Czech Technical University in Prague, Czech Republic}
\affil[3]{Université de Lille, CNRS, Centrale Lille, UMR 9189 CRIStAL, F-59000 Lille, France}
\affil[4]{National Research Council of Canada \& Systems Design Engineering, University of Waterloo, Canada}

\date{}

\maketitle

\begin{abstract}
Sign-language recognition has achieved substantial gains in classification accuracy in recent years; however, the latency and power requirements of most existing methods limit their suitability for real-time deployment. Neuromorphic sensing and processing offer an alternative paradigm based on sparse, event-driven computation that supports low-latency and energy-efficient perception.

In this work, we introduce an end-to-end neuromorphic architecture for American Sign Language (ASL) fingerspelling recognition that integrates a spiking visual attention mechanism for online region-of-interest extraction with a compact spiking neural network deployed on the SpiNNaker neuromorphic platform. We benchmark the proposed system against two datasets: a synthetically generated event-based version of the Sign Language MNIST dataset and a natively recorded ASL-DVS dataset, whilst providing a comprehensive overview of \textcolor{black}{Sign-language} recognition and related work.

This work yields competitive performance in simulation (92.27\%) and comparable performance on neuromorphic hardware deployment (83.1\%), while achieving the most energy-efficient architecture (0.565~mW) and low latency (3~ms) across all benchmarked approaches. Despite its compact design, the system demonstrates the suitability of task-dependent visual attention applications for edge deployment.

\end{abstract}



\section{Introduction}


Natural hand gestures, including Sign-language, play an important role in human communication, particularly for deaf and hard of hearing users.
Recent literature focuses primarily on accuracy of Sign-language recognition~\cite{li2023multimodal}; however, for interaction to be effective and reactive, gesture-recognition systems must achieve low latency, low power consumption, and high robustness to environmental variability. Latencies exceeding 100-200~ms break the illusion of agency~\cite{card1991information,miller1968response}, leading to user frustration and reduced trust. Consequently, Sign-language recognition in interactive settings requires perception and decision-making to be continuously responsive, requiring lightweight and power-efficient systems suitable for deployment on robotic platforms or wearable devices.

In humans, visual attention mechanisms enable selective processing of task-relevant regions of interest (ROIs), focusing perceptual resources toward salient scene elements while filtering irrelevant information~\cite{rizzolatti1983mechanisms}. This offers a bio-inspired strategy for achieving both efficiency and responsiveness. 
By contrast, modern, data-driven systems~\cite{gizdov2025seeing} that process the entire visual field uniformly allocate computational resources to non-informative regions of the scene, compromising both efficiency and robustness.

 \begin{figure}[htbp]
     \centering
     \includegraphics[width=0.5\textwidth]{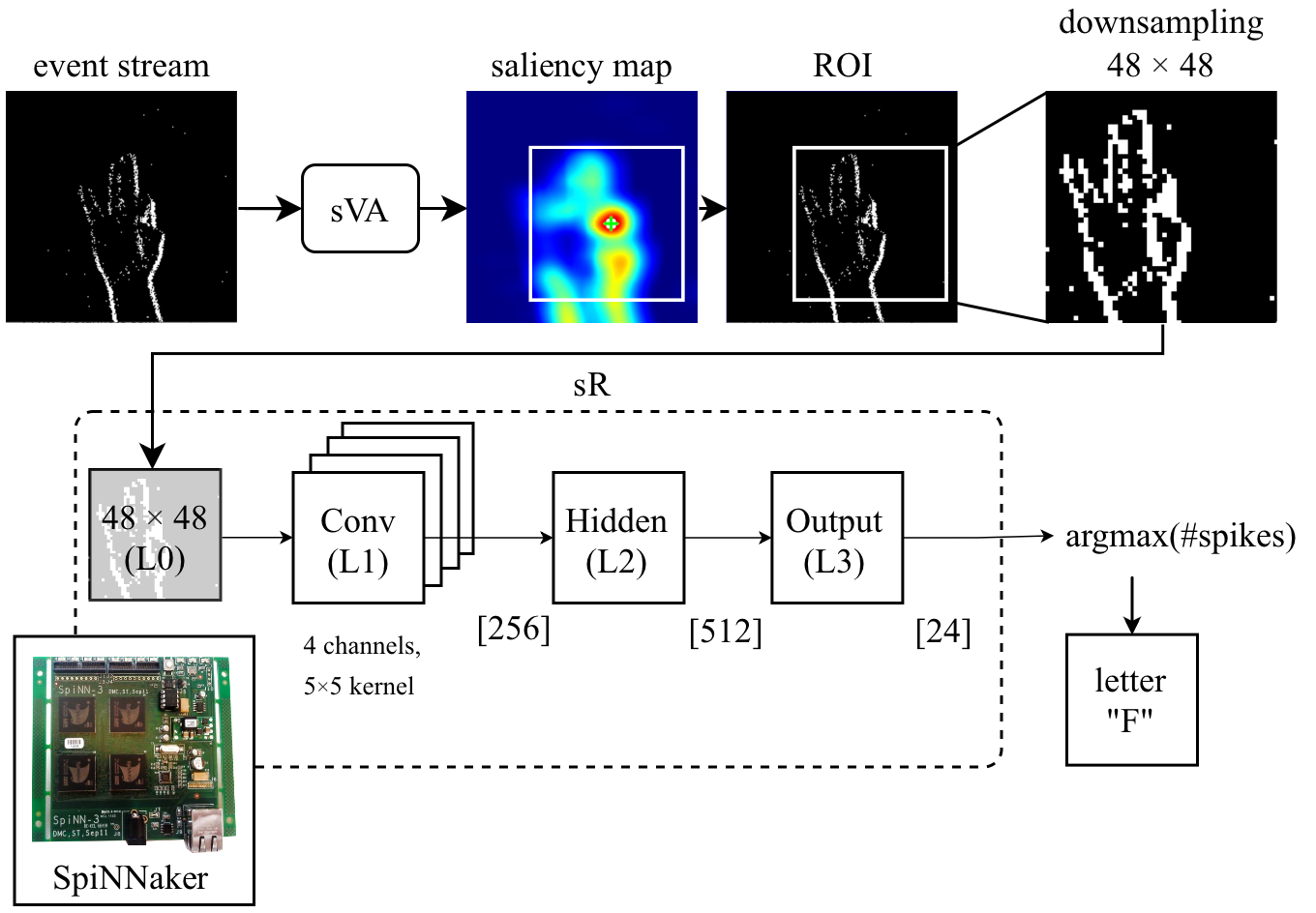}
     \caption{\textbf{Neuromorphic \textcolor{black}{Sign-language}  recognition architecture:} Event streams are processed through the spiking visual attention (sVA) model to extract salient regions (ROIs), which are then downsampled and fed into the recognition network (sR) deployed on SpiNNaker. The network architecture comprises convolutional (L1, 256 neurons, 4 channels, $5\times5$ kernel), hidden (L2, 512 neurons), and output (L3, 24 output neurons) layers.}
     \label{fig:system_diagram}
 \end{figure}

Without attentional selectivity, these approaches incur significant computational burdens~\cite{gizdov2025seeing} and exhibit weaknesses in generalisation across varying scales, orientations and illumination conditions~\cite{zhu2020review}.
Biologically inspired frame-based attention mechanisms~\cite{russell2014model,molin2013proto,itti1998model} can alleviate these limitations by reducing the volume of data requiring detailed analysis, focusing computational resources on ROIs that contain the relevant information in the scene.

However, these frame-based bio-inspired visual attention models~\cite{russell2014model,molin2013proto,itti1998model} rely on dense RGB frames, which are prone to motion blur, limited dynamic range, and high data throughput. By contrast, event-based or Dynamic Vision Sensors (DVS)~\cite{b4,gallego2020event} achieve efficiency at the hardware level through asynchronous, sparse encoding of visual changes.

Inspired by the mammalian retina, DVS devices asynchronously report per-pixel illumination changes with microsecond temporal resolution and signed polarity. This event-driven mechanism eliminates motion blur, increases invariance to illumination changes, extends dynamic range, and reduces latency. By transmitting only spatiotemporal changes rather than complete frames, the data throughput, power consumption, and storage requirements are reduced. These advantages make event-based cameras particularly suited for real-time applications.

Event-driven bio-inspired saliency-based visual attention models~\cite{iacono2019proto,ghosh2022event,d2025event} 
address the data acquisition challenges of conventional vision systems, achieving processing latencies of $\sim$100~ms for event-based attention~\cite{iacono2019proto}. 
However, realising their full potential requires a corresponding shift in computational processing, as frame-based processing of event data introduces a fundamental mismatch between asynchronous sensing and synchronous computation.
Neuromorphic hardware~\cite{douglas1995neuromorphic}, whether analog~\cite{pehle2022brainscales,moradi2017scalable,qiao2015reconfigurable} or digital~\cite{Furber2014,davies2018loihi,sga2020neuromorphic}, eliminates the von Neumann bottleneck by co-locating memory and processing, enabling Spiking Neural Networks (SNNs) to process asynchronous event streams in a massively parallel fashion~\cite{DiehlS2015SpikingDeepNetworks, 7838165,gehrig2020eventbasedangularvelocityregression,SpikeMS}, with activity-scaled power consumption mirroring biological neural efficiency, making them suited for real-time applications~\cite{fatahi2024event,d2020event}.

These gains are demonstrated concretely by event-based saliency models, 
which achieve a 6$\times$ latency reduction (from $\sim$100~ms~\cite{iacono2019proto} to $\sim$16~ms~\cite{Dangelo_etal22}) 
when migrated to SpiNNaker neuromorphic hardware.
Building on SNN saliency-based foundations~\cite{Dangelo_etal22}, D'Angelo et al.~\cite{d2025wandering} developed an end-to-end spiking-based, learning-free visual closed-loop exploration architecture on a pan-tilt unit, achieving competitive salient object detection across diverse lighting conditions with $\sim$124~ms latency using Metal Performance Shaders.

However, all the aforementioned works do not integrate attention mechanisms with concrete recognition tasks, limiting their applicability to goal-directed visual processing.

Gesture recognition represents one such task where this integration would be beneficial. Existing approaches process DVS data using either conventional artificial neural networks (ANNs)~\cite{rutishauser20237} or SNNs~\cite{chen2023sign,lee2014real,xiao2022dynamic}, often not deployed on dedicated neuromorphic hardware and instead relying on conventional CPUs or GPUs.
More recent work has demonstrated gesture recognition systems deployed directly on neuromorphic platforms~\cite{vasudevan2020introduction, ceolini2020hand, amir2017low, massa2020efficient, arfa2025efficient, nazeer2024language, stewart2020online}, exploiting the full potential of neuromorphic computing (See the final Table~\ref{tab:benchmark} for a detailed comparison).
Therefore, the majority of DVS datasets focus on dynamic gestures, as the motion information captured by event cameras provides advantages over conventional RGB cameras for tracking rapid movements. 

In contrast, few datasets address static signs and hand shapes, such as Sign-language fingerspelling alphabets, which constitute an important component of Sign-language communication. State-of-the-art fingerspelling recognition is dominated by frame-based RGB cameras and with deep ANNs, operating on images or video clips~\cite{b2}. Although RGB-based deep learning achieves 90–98\% accuracy on benchmarks of static fingerspelling or continuous signing sequences~\cite{b3}, it remains ill-suited to deployments where responsiveness and energy efficiency are as critical as recognition accuracy. 
To the best of the authors' knowledge, ASL-DVS~\cite{bi2020graph} is the only natively recorded DVS dataset focused on static fingerspelling, with recognition performed on a GPU; although there are natively recorded DVS datasets of dynamic Sign-language gestures available ~\cite{chen2023sign}.
The rest of the fingerspelling datasets consist of frame-based RGB datasets, such as the Sign Language MNIST (\acs{mnist})~\cite{SLmnist}, or the ASL fingerspelling Dataset~\cite{pugeault2011spelling}. Therefore, no existing end-to-end spiking-based fingerspelling recognition system leverages neuromorphic computing to fully exploit its inherent low-latency and low-power advantages. 

To address this gap, we propose an architecture for fingerspelling recognition in American Sign Language (ASL) that leverages end-to-end spiking attention-driven ROI selection to reduce input data, followed by fingerspelling classification on the neuromorphic platform SpiNNaker~\cite{Furber2014}, which minimizes computational loads through energy-efficient neuromorphic processing.
This work targets static fingerspelling, leveraging neuromorphic capabilities to capture individual letter poses before the transient event data decays, demonstrating ultra-low-latency efficiency for real-time applications.

\noindent The contributions of this work are listed below: 

\noindent \textit{1)} System-level integration of a lightweight, spiking-based, learning-free visual attention model~\cite{d2025wandering} that selectively allocates computational resources to task-relevant regions and effectively filters background clutter with a spiking-based fingerspelling recognition model.

\noindent \textit{2)} Design of an efficient and compact neuromorphic architecture for recognition deployed on the SpiNNaker neuromorphic platform~\cite{Furber2014} (SpiNN-3), explicitly constrained by the hardware resource limits (neurons, synapses, routing).

\noindent \textit{3)} Benchmark evaluation on two fingerspelling datasets: Sign Language MNIST~\cite{SLmnist} and ASL-DVS~\cite{bi2019graph},  where the former is converted to events using a standard pipeline and made available for reproducible benchmarking (Section~\ref{sec:code}).

\noindent \textit{4)} Validation through both simulation and hardware deployment on SpiNNaker.

\noindent \textit{5)} System characterization of latency and power consumption measured on the SpiNNaker platform.

\noindent \textit{6)} Comprehensive comparison of state-of-the-art gesture and fingerspelling recognition systems (final comparison Table~\ref{tab:benchmark}): accuracy, inference latency, and power consumption.

\section{Methods}
\subsection{spiking-based Visual Attention model}
This work (see Figure~\ref{fig:system_diagram}) extends the spiking-based visual attention model (\acs{att}) proposed by D'Angelo \emph{et al.}~\cite{d2025wandering,Dangelo_etal22} to process asynchronous event streams from event cameras. The \acs{att} model employs a bio-inspired saliency mechanism~\cite{russell2014model} to extract relevant bottom-up information from the visual field. This model takes inspiration from border ownership cells in the visual cortex~\cite{zhou2000coding}, which perceptually group information to identify salient regions that potentially contain objects. The model emulates this functionality by pooling the information from several orientations of the curved von Mises filter (see Equation~\ref{eq:VMeq}), where 
$x$ and $y$ are the kernel coordinates with the origin at the center of the filter, $\theta$ represents its orientation, $I_0$ is the modified Bessel function of the 
first kind, $R_0$ is the radius of the filter representing the distance of the VM  filter from the center, and $\tan^{-1}$ takes two arguments returning values in radians in the range $(-\pi, \pi)$.

\begin{equation} \label{eq:VMeq}
    VM_{\theta} (x,y) = \frac{\exp(\rho \cdot R_0 \cdot \cos(\tan^{-1}(-y, x) - \theta)}{I_0(\sqrt{x^2 + y^2 - R_0})}
\end{equation}

This radius will later define the radius of the opposite orientation of the von Mises filters, grouping information to detect the proto-object and defining the range of detectable sizes. Our approach employs a two-stage pipeline: the \acs{att} extracts regions of interest from event streams, which are then processed by a compact feed-forward spiking neural network (\acs{sr}) for Sign-language fingerspelling recognition.

\subsection{spiking Recognition model} \label{sec:sR}
The \acs{sr} model processes the extracted visual attention ROIs, operating on downsampled $48\times48$ event inputs, a resolution empirically determined to be sufficiently large to preserve relevant visual information while satisfying SpiNN-3 resource constraints.
The network architecture (see Table~\ref{tab:network_params}) consists of a single convolutional layer (L1) (4 channels, $5\times5$ kernel) of 256 neurons for spatial feature extraction, followed by a fully connected Hidden layer (L2) of 512 neurons and a linear readout Output layer (L3) with 24 output neurons representing the 24 static letters of the ASL fingerspelling alphabet, with L2 and L3 using LIF dynamics and classification determined by argmax over output spike counts.

The architecture is shaped by SpiNN-3 hardware 
constraints: The size of L3 is fixed by the number of classes, while L1 is required to spatially downsample the input to a tractable representation within SpiNNaker's connectivity limits by mapping the 48$\times$48 input into 4 channels, each represented by an 8$\times$8 feature map, yielding a total of 256 neurons, while the 5$\times$5 kernel size ensures spatial selectivity without incurring excessive synaptic connectivity. The size of L2 (512 neurons) is selected to maximally utilise the remaining neuron and synapse budget. The number of timesteps 
is likewise constrained by the real-time latency objective. The presented 
architecture reflects a hardware-feasible trade-off operating point consistent 
with the goal of demonstrating a real-time, on-edge neuromorphic system.


\begin{table}[htbp]
\centering
\caption{Network architecture and connectivity parameters.}
\label{tab:network_params}
\begin{tabular}{lcccc}
\toprule
\textbf{Population} & \textbf{\# Neurons} & \textbf{Input Syn.} & \textbf{Connection} \\
\midrule
Input (L0) & 2304 (48$\times$48) & --- & --- \\

Conv (L1) & 256 & 6400  & Conv.\\ 

Hidden (L2) & 512 & 130938 & Fully conn. \\

Output (L3) & 24 & 12284 & Fully conn. \\

\bottomrule
Total  & 3096 & 149622 &   &   \\
\end{tabular}
\end{table}

\section{Experiments \& Results}
\label{sec:experiments}

The system architecture is validated through simulation using snnTorch~\cite{snnTorch} and hardware deployment on SpiNNaker. Event streams are processed by the \acs{att} on a MacBook with an Apple M4 Pro processor, reserving SpiNNaker's computational resources entirely for the \acs{sr}.
Although the \acs{att} runs on a conventional processor in this work, the pipeline is 
fully spiking and 
neuromorphic-compatible: the lightweight \acs{att} model has been previously validated on SpiNNaker neuromorphic hardware for real-time operation~\cite{Dangelo_etal22}, and the current split reflects the neuron and synapse budget 
constraints of the SpiNN-3 board rather than an architectural limitation.
The \acs{att} model achieves 88.8\% accuracy for salient object detection (against ground truth segmentation masks) in office scenarios and 89.8\% in challenging indoor and outdoor low-light conditions~\cite{d2025wandering} on the event-assisted low-light video object segmentation dataset~\cite{li2024event}.
The ROI extracted by \acs{att} is then fed into the \acs{sr} model for the sign-letter classification. We tested the \acs{sr} model in two conditions: in simulation on an Apple processor using the Metal Performance Shaders, and the other running the spiking model directly on the SpiNNaker. 

\subsection{Event-based Sign-language datasets}

The event-based vision community has predominantly focused on gesture recognition~\cite{rutishauser20237,arfa2025efficient,ceolini2020hand}, providing many DVS dynamic gesture datasets, such as the IBM DVSGesture Dataset~\cite{amir2017low} or the SL-Animals-DVS Dataset~\cite{vasudevan2020introduction}, but the availability of native DVS datasets for Sign-Language fingerspelling (static signs) remains limited. 
To validate the \acs{sr} model, we evaluate our approach on two ASL fingerspelling datasets: Sign language MNIST dataset (\acs{mnist})~\cite{SLmnist} and the ASL-DVS~\cite{bi2019graph}. Sign Language MNIST allows comparison with existing SNN architectures deployed on Loihi~\cite{mohammadi2022static} using the same benchmark, while ASL-DVS is, to the best of our knowledge, the only natively event-based fingerspelling dataset available. 

\vspace{.8em}
\subsubsection{Sign Language MNIST}
The Sign Language MNIST dataset (\acs{mnist})~\cite{SLmnist} is a widely-used benchmark for ASL fingerspelling recognition, consisting of 34627 grayscale $28\times28$ images of 24 static letters (J and Z excluded because of dynamic nature of the gestures), divided into 27455 training and 7172 test samples. Preprocessing included cropping, grayscale conversion, resizing, and data augmentation via filtering, pixelation, and rotation. The hand isolation step introduces computational overhead incompatible with real-time low-latency operation.

To test our architecture, events are generated by applying a random walk pixel shift to the static grayscale images, creating synthetic video sequences that are subsequently converted to event streams using the IEBCS (Image-to-Event-Based-Camera Simulator) framework~\cite{IEBCS}. 
The conversion was performed once with a fixed random seed, producing a static event-based dataset with no run-to-run variance. These event streams are processed by the lightweight \acs{att} model to extract ROIs in a computationally efficient manner suitable for online demonstration, as previously shown by D'Angelo \emph{et al.}~\cite{d2025wandering}. For experimental consistency, we adopted the original train–test split defined by the authors of the \acs{mnist} dataset.
 
\vspace{.8em}
\subsubsection{ASL-DVS dataset}
The ASL-DVS dataset (\acs{asl}), introduced by Bi et al.~\cite{bi2019graph}, is a large natively recorded event-based dataset for Sign-language fingerspelling recognition under real-world conditions. The dataset was acquired using an iniLabs DAVIS240c neuromorphic vision sensor positioned in an office environment with low environmental noise and constant illumination. It contains over 100000 samples across 24 classes, like Sign Language MNIST; it excludes letters J and Z. Five subjects were recorded performing the different static handshapes relative to the camera to introduce natural variance into the dataset. For each letter class, 4200 samples were collected, with each sample capturing approximately 100~ms of asynchronous event data. Following this subject-wise organization, we adopted a cross-subject evaluation protocol, using data from four signers for training and validation and reserving the remaining signer exclusively for testing.

\subsection{Role of spiking-based Visual Attention}
\begin{figure}[htbp]
  \centering
  \includegraphics[width=0.47\textwidth]{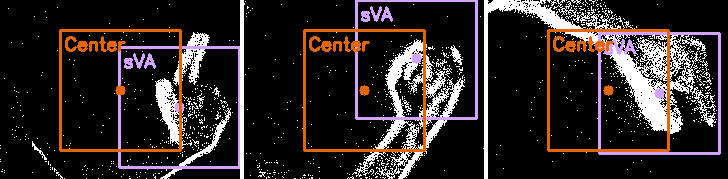}
  \caption{\textbf{sVA ROI (purple) vs.\ naive center crop (orange)} on three ASL-DVS frames (letters L, O, Q).}
  \label{fig:sva_ablation}
\end{figure}

Figure~\ref{fig:sva_ablation} illustrates the need for the \acs{att} module on ROI selection, contrasted against a naive center-crop baseline. Since subjects vary their hand position freely, center-cropping frequently captures background rather than the hand, while \acs{att} reliably localises the salient region regardless of position.

\subsection{Simulation}
The \acs{sr} model, described in Table~\ref{tab:network_params}, is implemented using snnTorch~\cite{snnTorch} and simulated on a MacBook with an Apple M4 Pro processor. All neurons of L2 \& L3 follow LIF dynamics with fixed membrane decay $\beta=0.92$ (See parameters in Table~\ref{tab:training_params}). For both datasets, each input sample was simulated for 35 discrete timesteps to align with the temporal extent of the event-based window data, as in~\cite{bi2019graph}. Additionally, given the heavy downsampling of the original \acs{mnist} dataset, we explored an extended timestep of 100~ms to investigate whether prolonged temporal integration could improve model performance.

\begin{table}[htbp]
\centering
\caption{Training configuration on snnTorch.}
\label{tab:training_params}
\small
\begin{tabular}{lcc c}
\hline
\textbf{Parameter} 
& \multicolumn{2}{c}{\textbf{\protect\acs{mnist}}} 
& \textbf{\protect\acs{asl}} \\
\hline
Leak factor ($\beta$) & 0.92 & 0.92 & 0.92 \\
Time steps            & 100   & 35   & 35   \\
Epochs                & 310  & 130  & 170  \\
\hline
\end{tabular}
\end{table}

The network is trained end-to-end via surrogate gradient backpropagation (See Table~\ref{tab:optim_params}) through time with a fast-sigmoid surrogate function, using AdamW optimization (learning rate $10^{-3}$, weight decay $10^{-4}$), Kaiming initialisation for convolutional and Xavier initialization for fully connected layers. To address class imbalance and improve performance on hard-to-classify samples (such as letter "U" against "V" and "M" against "N", etc.), training employed focal loss with label smoothing ($\epsilon$), class-balanced weighting, and online hard example mining. The learning rate followed a cosine annealing schedule with warm restarts to improve convergence stability (See Table~\ref{tab:optim_params}).

\begin{table}[htbp]
\centering
\caption{Training optimization parameters on snnTorch.}
\label{tab:optim_params}
\small
\begin{tabular}{p{0.18\columnwidth}p{0.72\columnwidth}}
\toprule
\textbf{Component} & \textbf{Parameters} \\
\midrule
Loss & Focal: $\gamma=2.0$, $\varepsilon=0.1$, mining thr: $0.65$\\
Optimizer & AdamW: lr $3\times10^{-3}$, $\beta=(0.9,0.999)$, decay $10^{-4}$\\
Scheduler & Cosine: $T_0=25$, $T_{\text{mult}}=2$, $\eta_{\min}=2\times10^{-7}$  \\
\bottomrule
\end{tabular}
\end{table}

Classification is performed by temporally accumulating output spikes and selecting the class with maximal spike count. The bias-free, compact architecture enables direct deployment on neuromorphic hardware platforms such as SpiNNaker. 

 \begin{figure}[htbp]
     \centering
     \includegraphics[width=0.48\textwidth]{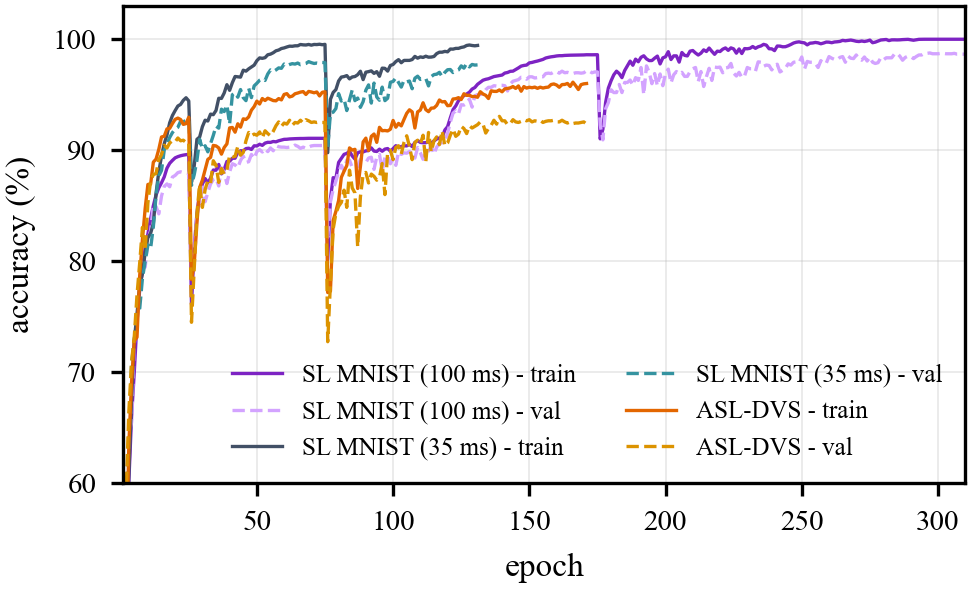}
     \caption{\textbf{Training and validation accuracy curves} for \acs{mnist} dataset (35~ms and 100~ms samples) and for the \acs{asl} dataset (35~ms samples).}
     \label{fig:training_curves}
 \end{figure}

Figure~\ref{fig:training_curves} shows the learning dynamics across both datasets. In both cases, the \acs{sr} model exhibits rapid initial learning, demonstrating efficient adaptation to event-based Sign-language fingerspelling data. The \acs{mnist} dataset shows faster initial convergence compared to \acs{asl}, likely due to its synthetic generation via random-walk pixel shifts, whereas \acs{asl} captures natural human motion.
Training and validation curves indicate consistent learning without significant overfitting.
Table~\ref{tab:sim_test_accuracy} reports test set performance evaluated in simulation. \acs{mnist} achieves 74.61\% accuracy for 35~ms and 83.82\% accuracy for 100~ms, while \acs{asl} reaches 92.27\%, both showing performance gaps compared to validation results in Figure~\ref{fig:training_curves}.
The validation–test gap on \acs{mnist} is attributed to stochastic variability introduced by the random-walk–based RGB-to-event conversion, which affects event timing and sparsity despite identical preprocessing across train, validation and test splits. Increasing the temporal integration window from 35~ms to 100~ms partially alleviates this effect by enabling additional temporal evidence accumulation, indicating that event sparsity and temporal variability are key contributors to the reduced test accuracy on \acs{mnist}.
By contrast, the natively recorded \acs{asl} dataset captures real human hand motion with real DVS sensors, resulting in richer event patterns and a smaller generalization gap and higher test accuracy.
Additional differences between the datasets, including the lower spatial resolution of the original \acs{mnist} images and the upsampling applied prior to event conversion, may further contribute to the observed performance disparity relative to natively recorded DVS data.
In the original work by Bi et al.~\cite{bi2020graph}, the event-based \acs{asl} dataset is addressed using graph-based event representations and graph and residual-graph CNN architectures. In contrast, we adopt a compact spiking network consisting of a single convolutional layer followed by fully connected stages. Despite its reduced architectural complexity, our model achieves an accuracy of 92.2\%, which is improved compared to the reported performance of the G-CNN (87.5\%) and RG-CNN (90.1\%). These results indicate that competitive recognition performance can be obtained using a compact spiking architecture. 


\subsection{SpiNNaker deployment}
The \acs{sr} model was then deployed on a SpiNN-3~\cite{Furber2014} development board containing 4 SpiNNaker chips, each equipped with 18 ARM968E-S cores. The membrane time constant $\tau_m$ is derived from the leak factor $\beta$ using the relationship $\tau_m = -\Delta t / \ln(\beta)$. Excitatory and inhibitory synaptic time constants ($\tau_{\text{syn}}^E$, $\tau_{\text{syn}}^I$) control synaptic memory dynamics, while the refractory period $\tau_{\text{refrac}}$ provides burst suppression. Weight scaling factors ($w_{\text{fc}}$, $w_{\text{out}}$) and inhibitory weights ($w_{\text{inh}}$) modulate connection strengths between layers (See Table ~\ref{tab:spinnparam} for the parameters of the deployed network).


\begin{table}[htbp]
\centering
\caption{SpiNNaker deployment parameters.}
\label{tab:spinnparam}
\small
\begin{tabular}{l c | l c}
\hline
\textbf{Parameter} & \textbf{Value} & \textbf{Parameter} & \textbf{Value} \\
\hline
$\Delta t$ (ms) & 1.0 
& $\tau_{\text{syn}}^E$ (ms) & 5.0 \\

$d$ (ms) & 1.0 
& $\tau_{\text{syn}}^I$ (ms) & 3.0 \\

$\tau_m$ (ms) & $-\Delta t/\ln(\beta)$
& $\tau_{\text{refrac}}$ (ms) & 1.0 \\

$V_{\text{rest}}$ (mV) & -65
& $w_{\text{fc}}$ & 0.3 \\

$V_{\text{reset}}$ (mV) & -65
& $w_{\text{out}}$ & 2.0 \\

$V_{\text{thresh}}$ (mV) & -61
& $w_{\text{inh}}$ & 1.0 \\
\hline
\end{tabular}
\end{table}

\subsubsection{\textcolor{black}{Accuracy}}
Table~\ref{tab:sim_test_accuracy} shows the accuracy values on SpiNNaker for the \acs{mnist} and the \acs{asl} dataset as 71.7\% and 83.1\%, respectively. Only the 100~ms \acs{mnist} model was deployed on hardware, for two reasons: it enables a fair comparison with \acs{asl}, and the 35~ms simulation with an accuracy of 74.61\% is too low to yield a meaningful hardware result given the known simulation-to-hardware gap.
The observed performance gap between simulation and deployment is primarily attributed to weight quantization:  SpiNNaker uses fixed-point weight representation versus the 32-bit floating-point precision of snnTorch, reducing the effective resolution of synaptic strengths. Additional factors include differences in synaptic dynamics and the more biologically realistic LIF formulation employed on SpiNNaker. Quantization-aware training and precision-aware calibration~\cite{vanAlbada2018} are planned for future work to reduce this discrepancy. Despite these differences, the overall behavior of the deployed network closely follows the simulation results, indicating that the trained model transfers effectively to the neuromorphic hardware platform.

\vspace{-2mm}
\begin{table}[htbp]
\centering
\caption{\acs{sr} test set performance in simulation and on SpiNNaker.}
\label{tab:sim_test_accuracy}
\begin{tabular}{l|cc|c}
\toprule
 & \multicolumn{2}{c|}{\textbf{Accuracy \acs{mnist}}} & \textbf{Accuracy \acs{asl}} \\
\textbf{Platform} & 35 ms & 100 ms & 35 ms \\
\midrule
Mac (simulation) & 74.61\% & 83.82\% & 92.27\% \\
SpiNNaker        & --      & 71.7\% &  83.1\%\\
\bottomrule
\end{tabular}
\end{table}

\vspace{-2mm}

\begin{figure}[htbp]
  \centering
 \includegraphics[width=0.47\textwidth]{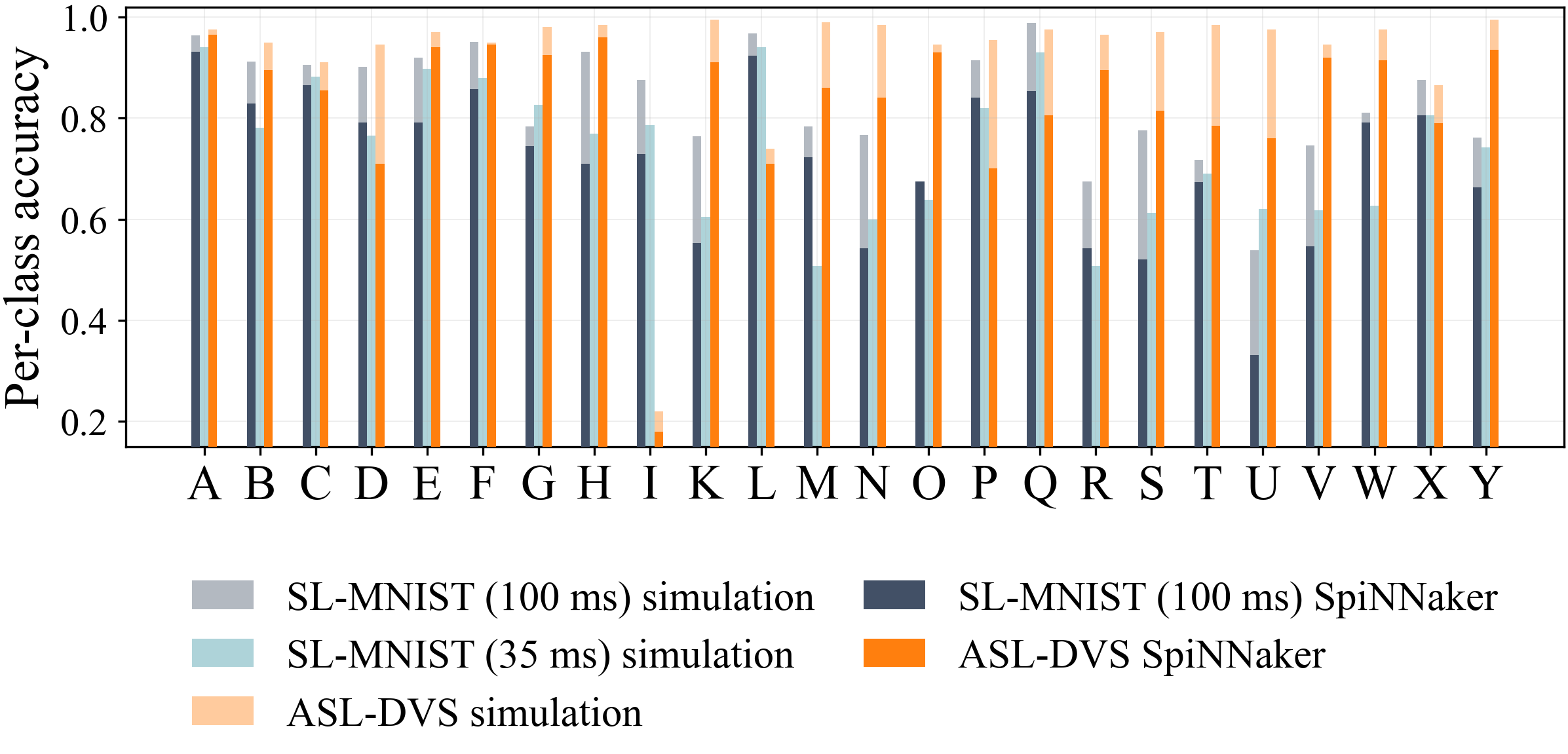}
 \caption{\textcolor{black}{\textbf{Comparison of per-class test accuracy of the \acs{sr} model evaluated in simulation and deployed on SpiNNaker} for the \acs{mnist} and the \acs{asl} dataset.  The horizontal axis indicates the ASL fingerspelling classes, while the vertical axis reports classification accuracy (\%).}}     
 \label{fig:spinn_per_class_accuracy}
\end{figure}

Figure~\ref{fig:spinn_per_class_accuracy} shows the per-class simulation and after-deployment accuracy for each letter of the ASL alphabet, demonstrating consistent performance across the native event-based \acs{asl} dataset, with the exception of letter "I", which exhibits notably lower accuracy likely due to visual similarity with other signs. Both simulated and deployed models show consistent performance between 35 and 100~ms temporal windows of \acs{mnist} dataset samples, with the longer duration yielding superior results, indicating that the model benefits from extended temporal integration.

\vspace{.8em}

\subsubsection{Latency and Power consumption}\label{sec:power_consumption}

To demonstrate the advantages of neuromorphic hardware, we report the latency and power consumption of the network deployed on the SpiNNaker platform (See Table~\ref{tab:benchmark} for final comparison). The network architecture comprises two sequential layers: input-to-hidden and hidden-to-output connections. Given SpiNNaker's fixed internal digital clock of $1~ms$, the total inference latency is $3~ms$ (one time step per layer plus input processing).

Following standard SpiNNaker power models~\cite{stromatias2013power, stromatias2015scalable}, total power is given by $P_T = P_I + P_B + (P_N \times n) + (P_S \times s)$, where $P_I$ is idle power, $P_B$ is baseline power for system services, $P_N$ is per-neuron power, and $P_S \approx 8$~nJ is the energy per synaptic event. 
We focus on \emph{dynamic power} ($P_S \times s$) to quantify neuromorphic processing efficiency independent of system overhead. Table~\ref{tab:spikes_per_layer} presents the average number of synaptic events per layer, calculated by dividing the total spike count by 120 samples, where each sample comprises the events collected over a 35 timesteps (where each $timestep=1~ms$) for each Sign-language gesture. The estimated energy consumption is approximately 24891.84~nJ and 19807.68~nJ, corresponding to a power consumption of approximately 0.711~mW for the \acs{mnist} and 0.565~mW for the \acs{asl} dataset, as shown in Table~\ref{tab:spikes_per_layer}. sVA latency, validated on MPS in~\cite{d2025wandering}, is 68.29~ms $\pm$ 6.19~ms on \acs{asl} dataset (M4 MPS).

\begin{table}[h]
\centering
\caption{Power consumption measurements}
\label{tab:spikes_per_layer}
\begin{tabular}{lrr}
\toprule
\textbf{Layer} & \textbf{\acs{mnist} Avg Spikes} & \textbf{\acs{asl} Avg Spikes} \\
\midrule
Conv (L1) & 719.31 & 720.95\\
Hidden (L2) & 2235.82 & 1681.92 \\
Output (L3) & 156.35 & 73.09 \\
\midrule
Total & 3111.48 & 2475.96 \\
Energy consumption & $\approx$~24891.84~nJ & $\approx$~19807.68~nJ\\
Power consumption & $\approx$~0.711~mW & $\approx$~0.565~mW\\
\bottomrule
\end{tabular}
\end{table}

\begin{table*}[!ht]
    \centering
    \begin{tabular}{c|p{2.3cm}p{1.8cm}cp{1.75cm}p{1.75cm}p{.5cm}cp{2cm}}
        \textbf{Ref} & \textbf{Dynamic Gestures} & \textbf{ASL}	& \textbf{Sensor} &\textbf{Network} & \textbf{Hardware} &	\textbf{Acc.} &	\textbf{Latency} & \textbf{Power}\\
        & & \textbf{Fingerspelling} & & & & & \textbf{(ms)} & \textbf{Usage (mW)}\\
        \toprule
         
        \cite{chen2023sign} & DVS\_Sign& - & DVS & SNN & CPU, GPU & 77\% & - & -\\
        
        \cite{lee2014real} & 11 Kinect gestures & - & stereo DVS & SNN + HMM & CPU & $>$90\% & - & -\\ 
        
        \cite{xiao2022dynamic} & DVSGesture IBM   & - &  DVS & LSM + CNN & CPU + GPU & \textbf{98.42\%} & - & - \\
        
        \cite{rutishauser20237} & DVSGesture IBM & - & DVS & TCN ANN & Kraken PULP SoC, CUTIE accelerator & 97.7\% & \textbf{0.9} & 4.7\\

         \midrule
         \cite{vasudevan2020introduction} & Gestures for ASL animals (19 signs) & - & DVS & SNN (SLAYER, STBP) & TrueNorth & - & - & - \\

         \cite{amir2017low} & DVSGesture IBM& - & DVS & conv SNN & TrueNorth & 96.5\% & \textbf{105} avg & 178.8\\

          \cite{ceolini2020hand} & 5 gestures + EMG & - & DVS + EMG & SNN & Intel Loihi & - & - & -\\

         \cite{massa2020efficient} & DVSGesture IBM & - & DVS & DNN-trained SNN & Intel Loihi & 89.64\% & 161.42 & -\\

         \cite{stewart2020online} & DVSGesture IBM & - & DVS & SNN & Intel Loihi & - & - &  -\\

        \cite{arfa2025efficient} &DVSGesture IBM & - & DVS & SNN & SpiNNaker2 & 94\% & - & 459 mJ\\
        
        \cite{nazeer2024language} & DVSGesture IBM  & - & DVS & EGRU SNN& SpiNNaker2 & 	\textbf{96.83\%} & - & 390\\
        \midrule

        \cite{rivera2017american} & - & 24 ASL Signs & DVS & ANN & FPGA & 79.58\% & - & -\\

        \cite{mohammadi2022static} & - & \acs{mnist}  & rate-based encoded RGB & SNN VGGNet  & Intel Loihi & \textbf{97.7\%} & 8.03 & 71.89\\
         &  &  & & SNN MLP &  & 90.69\% & 12.64 & 61.37\\
        
        \cite{bi2020graph} & - & \acs{asl} & DVS & G-CNN & GPU & 87.5\% & - & -\\
         &  &  & & RG-CNN & & 90.1\% & &\\

         \midrule
        \acs{sr} (this work) & - & \acs{mnist} & RGBtoDVS & SNN & SpiNNaker~\cite{Furber2014} & 71.7\%  & \textbf{3} & \textbf{0.711} \\

        \acs{sr} (this work) & - & \acs{asl} & DVS & SNN & SpiNNaker~\cite{Furber2014} & 83.1\% & \textbf{3} & \textbf{0.565}
        
    \end{tabular}
    \caption{Summary of recent work in DVS gesture recognition. Abbreviations: Ternarized Convolution Network (TCN), Hidden Markov Model (HMM), Artificial Neural Network (ANN), Spiking Neural Network (SNN), Liquid State Machine (LSM), Deep Neural Network (DNN), Event-based Gated Recurrent Unit (EGRU).}
    \label{tab:benchmark}

\end{table*}

\section{Benchmark}

Table~\ref{tab:benchmark} summarizes recent work on gesture recognition and Sign-language fingerspelling recognition, with emphasis on modality (DVS/RGB), hardware platform, and reported performance metrics. To provide an overview of research in this specific application domain, we also include gesture recognition studies.
While a substantial body of work has addressed dynamic gesture recognition using DVS data, and some models have been deployed on neuromorphic hardware such as TrueNorth~\cite{vasudevan2020introduction, amir2017low}, Loihi~\cite{ceolini2020hand, massa2020efficient, stewart2020online}, or SpiNNaker~\cite{arfa2025efficient, nazeer2024language}, only a small subset of studies focus on and report inference latency and power consumption~\cite{amir2017low,massa2020efficient,arfa2025efficient, nazeer2024language}.
The overall accuracy across dynamic gesture recognition models is reported to be typically high, exceeding 90\% in most cases. However, the majority of reported (or absent) latency and power consumption values in Table~\ref{tab:benchmark} suggest that classification accuracy is predominantly prioritized, with comparatively less emphasis on end-to-end efficiency metrics. 

In contrast, neuromorphic work focusing specifically on Sign-language fingerspelling recognition is scarce. Existing SNN-based approaches for sign and gesture recognition typically employ multiple convolutional layers followed by large fully connected stages, often comprising several hundred hidden units~\cite{ceolini2020hand,chen2023sign,vasudevan2020introduction}. While such architectures can achieve high classification accuracy~\cite{xiao2022dynamic,amir2017low}, their depth and parameter count pose challenges for extremely low-latency and energy efficiency.

A comparable architecture in terms of model compactness is the multilayer perceptron (MLP) proposed by Mohammadi et al.\cite{mohammadi2022static}, consisting of 1576 neurons, which was trained for \acs{mnist} recognition and deployed on the Intel Loihi platform. While this model achieves higher classification accuracy (90.69\%), it operates at a substantially higher inference latency (12.64~ms) and dynamic power consumption (61.37~mW). 
In contrast, our proposed \acs{sr} model achieves lower accuracy on the same \acs{mnist} benchmark, but reduces inference latency to 3~ms, 4x times reduced and dynamic power consumption 86x times. This comparison highlights the different operating points of the two approaches and illustrates the emphasis of the proposed system on ultra-low-latency and energy-efficient deployment.
The lightweight \acs{sr} model proposed in this work reports the lowest power consumption among existing Sign-Language recognition systems. On the \acs{asl} real-world DVS data benchmark, it achieves slightly lower but comparable accuracy in hardware deployment, whilst maintaining performance parity in simulation, despite its extremely compact architecture.
Moreover, the fully event-driven pipeline supports online, continuous inference with minimal latency, establishing a foundation for future real-time applications beyond the scope of this work.

\section{Code and Data Availability}\label{sec:code}
The open-source architecture is available at \footnote{\url{https://github.com/Neuro-inspired-Perception-and-Cognition/Sign-language-Recognition}}. Data of the event-based ~\acs{mnist} can be found here \footnote{\url{https://drive.google.com/drive/folders/1ti0ou-iGFHyBpfTag3ftHcD25nXTtrfj?usp=sharing}}.

\section{Conclusion}

This work proposes an end-to-end neuromorphic spiking-based pipeline that couples spiking visual attention with low-latency on-chip fingerspelling recognition on SpiNNaker~\cite{Furber2014} neuromorphic platform, demonstrating that effective ASL fingerspelling recognition is achievable under strict hardware resource constraints. 
Across both benchmark datasets~\cite{SLmnist,bi2020graph}, the system reports the lowest power consumption and among the lowest latencies of all compared approaches, while achieving competitive accuracy in simulation and comparable, though slightly reduced, accuracy in hardware deployment.

The results demonstrate that competitive recognition performance can be achieved with 
substantially reduced architectural complexity compared to prior SNN-based gesture and sign recognition approaches, while explicitly optimizing the end-to-end system for deployability on neuromorphic hardware.

The combination of real-time attention-driven ROI selection, high processing speed, low power consumption, and competitive accuracy suggests the potential of deploying this compact network in wearable devices and edge computing applications.

\vspace{-5mm}
\section*{Acknowledgments}
GDA was supported by the European Union’s HORIZON-MSCA-2023-PF-01-01 research and innovation programme under the Marie Skłodowska-Curie grant agreement ENDEAVOR No. 101149664. SL and MH were supported by the project ROBOPROX (reg. no. CZ.02.01.01/00/22\_008/0004590). SL and MH were additionally supported by the Czech Science Foundation (GAČR) project PIONEER, no. 26-23432S.
We thank Bedrich Himmel for technical assistance.

\bibliographystyle{ieeetr}
\bibliography{references.bib}

\end{document}